\documentclass[10pt,twocolumn,letterpaper]{article}

\usepackage{cvpr}
\usepackage{times}
\usepackage{epsfig}
\usepackage[pagebackref=true,breaklinks=true,letterpaper=true,colorlinks,bookmarks=false]{hyperref}

\usepackage{graphicx}
\usepackage{amsmath}
\usepackage{amssymb}
\usepackage{subcaption}
\usepackage{enumitem}
\usepackage{cleveref}

\usepackage{simplemargins}
\setbottommargin{0.55in}

\usepackage[pagebackref=true,breaklinks=true,letterpaper=true,colorlinks,bookmarks=false]{hyperref}

\cvprfinalcopy 

\def\cvprPaperID{21} 

\ifcvprfinal\fi
\begin{document}

\title{Instant 3D Object Tracking with Applications in Augmented Reality}

\author{Adel Ahmadyan \thanks{Equal contribution} \and Tingbo Hou \footnotemark[1] \and Jianing Wei \footnotemark[1]  \and Liangkai Zhang  \and Artsiom Ablavatski \and Matthias Grundmann,\\
Google Research\\
1600 Amphitheatre Pkwy, Mountain View, CA 94043\\
{\tt\small \{ahmadyan, tingbo, jianingwei, liangkai, artsiom, grundman\}@google.com}
}

\maketitle

\begin{abstract}
Tracking object poses in 3D is a crucial building block for Augmented Reality applications. 
We propose an instant motion tracking system that tracks an object's pose in space (represented by its 3D bounding box) in real-time on mobile devices. 
Our system does not require any prior sensory calibration or initialization to function. We employ a deep neural network to detect objects and estimate their initial 3D pose. Then the estimated pose is tracked using a robust planar tracker. 
Our tracker is capable of performing relative-scale 9-DoF tracking in real-time on mobile devices. 
By combining use of CPU and GPU efficiently, we achieve 26-FPS+ performance on mobile devices.
\end{abstract}
   
\section{Introduction}
Tracking in monocular videos is a challenging and well studied problem in computer vision. While 2D tracking is mature with robust solutions~\cite{Wei:2019,prince:02,Baker:2004,Benhimane:2004,Pirchheim:2011,Liang:2018}, 3D tracking from monocular RGB images remains an open problem. Current approaches to 3D tracking~\cite{hu2018joint,weng2019baseline,zhou2020tracking} require complex initialization procedures to estimate depth, and are not very robust and have high computational cost.

The objective of 3D tracking is to track the 3D bounding box of a rigid object throughout frames when both camera and object motions are present. The object pose in 3D is uniquely determined by its 3D bounding box and has 9 DoF including orientation, translation, and physical size. 

In this paper, we propose a system to detect and track an object's 3D pose in real-time. Initially, we detect the object and estimate its pose using a deep neural network~\cite{Hou:2020}. 
The detection network does not require prior knowledge of the object's shape, size or CAD model to be known and can detect category-level unseen objects.
This model can run in real-time on mobile devices and can be used in a \emph{tracking-by-detection} paradigm. When the model is applied to every frame, the detection output may suffer from jitter due to the prediction noise from the model. This jitter is undesirable for AR applications. We adopt a detection-plus-tracking framework to mitigate this issue. This framework mitigates the need to run the network on every frame, allowing the use of heavier and therefore more accurate models, while keeping the pipeline real-time on mobile devices. It also retains object identity across frames and ensures that the prediction is temporally consistent, effectively reducing the jitter.

Our detection-plus-tracking system works as follows. We first locate the object's pose using the detection network, estimating its 3D bounding box. We then project the bounding box's 3D vertices to the image plane and track the 2D points, which rest on a plane, using a planar tracker~\cite{Wei:2019}. Finally, we lift the tracked 2D points to 3D using the EPnP~\cite{Lepetit_2009_EPnP} algorithm to estimate the 3D bounding box in subsequent frames. This work extends our previous works on 3D object detection~\cite{Hou:2020}, instant motion tracking~\cite{Wei:2019}, and 3D object tracking~\cite{ahmadyan:2020}.  

Our tracker has three properties: it is robust, instant, and real-time. 
The tracker is very efficient in utilizing both CPU and GPU on-device for tracking and detection, respectively, thus achieving real-time (26-FPS+) performance on mobile devices. 
The detection is performed on a single RGB image and the planar tracker is instant. Thus our whole pipeline does not require any parallax-inducing motion to initialize and locate the object and its overall latency is low. Finally, with the assumption that 3D bounding box sits on a planar ground, the planar tracker is very robust given typical AR applications.

Our main contributions are:
\begin{itemize}
\setlength{\parskip}{0pt}
\setlength{\itemsep}{0pt plus 1pt}
    \item We propose an end-to-end system to track the object's 3D pose (orientation, translation, and size up to a scale). This system uses a CNN to initialize the pose, then track it using a planar tracker.
    \item The proposed system is calibration-free and does not require any complex initialization sequence or any hardware beyond camera or IMU sensors. It does not require prior knowledge of object's shape or model and can detect and track unseen objects.
    \item The end-to-end system (including detection) runs in real-time on mobile devices. 
\end{itemize}

To encourage researchers and developers to experiment and prototype based on our pipeline, we open-sourced our on-device ML pipeline in~\cite{mediapipe}, including an end-to-end demo mobile application and our trained detector for shoes and chairs. We hope that sharing our solution with the wide research and development community will stimulate new use cases, applications, and research efforts.

\section{Related work}
Efficient and robust tracking is an essential component for any AR application. Object tracking has been studied and practiced extensively in computer vision. Planar and region-based trackers \cite{prince:02,Baker:2004,Benhimane:2004,Pirchheim:2011,Liang:2018} rely on 3D geometry to estimate the camera motion and track objects. 
Recently, neural nets have been utilized to learn and estimate motion. Generally tracking system consists of two components: \\
a) detector: which detects the objects in each frame and estimates their 2D or 3D bounding boxes, and \\
b) a matching algorithm, which tracks object correspondences between frames.

\cite{hu2018joint} detects an object's 3D bounding box and then estimates the 3D motion between frames to track the box. Employing a Kalman filter after 3D detection was investigated in~\cite{weng2019baseline} and shown to achieve good performance. In \cite{li2018stereo} and \cite{osep2018combined}, tracking the 3D bounding box with stereo cameras for autonomy applications were studied. \cite{sharma2018pixels} used 3D cues for tracking vehicles' 2D bounding boxes. Recently, in~\cite{zhou2020tracking} the authors propose to use a deep neural network for both 3D detection and tracking.

\section{Instant 3D tracking}
\begin{figure}
\begin{center}
   \includegraphics[width=7cm]{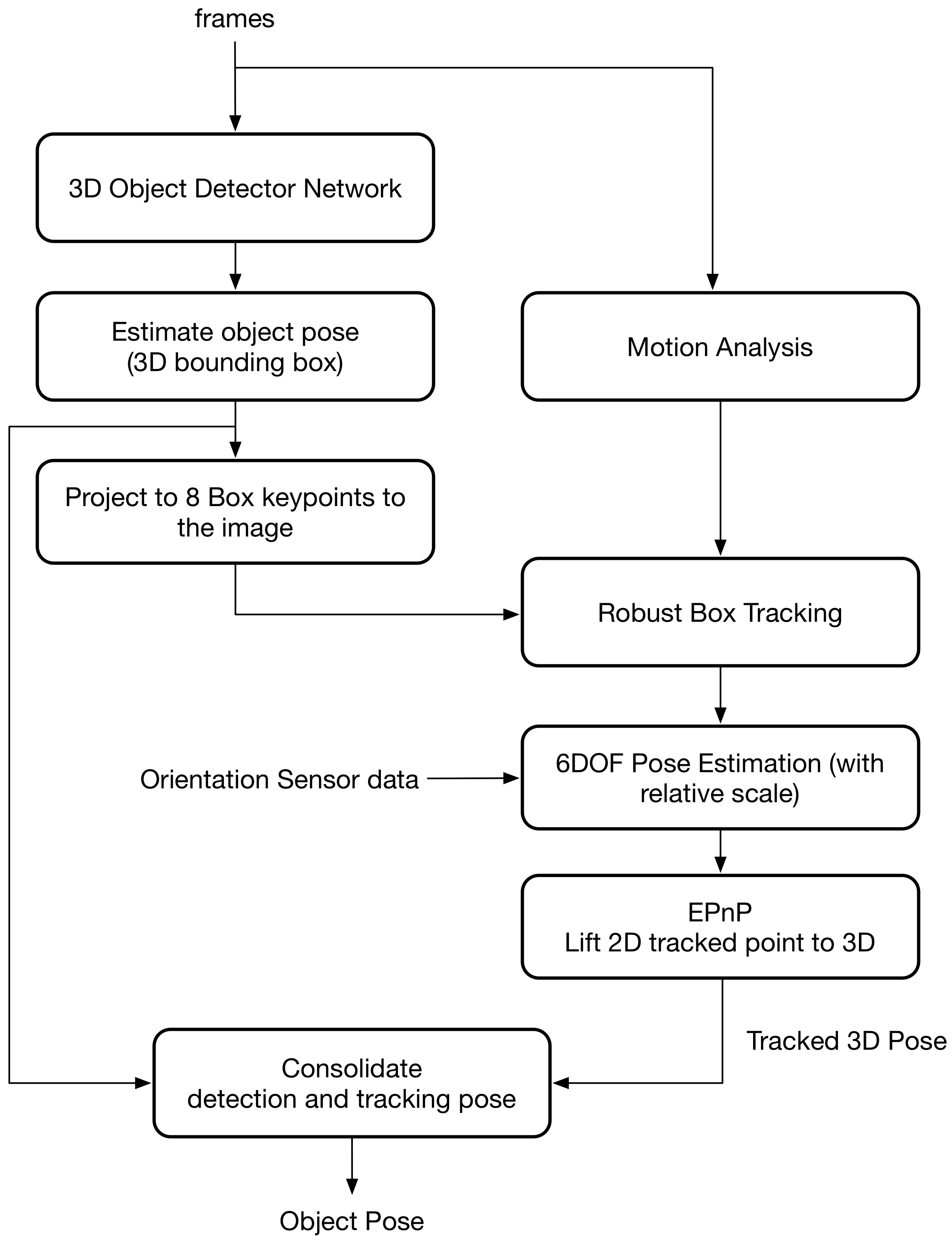}
\end{center}
   \caption{Overview of our 3D tracking system.}
\label{fig:overview}
\end{figure}
\begin{figure}
\begin{center}
   \includegraphics[width=8cm]{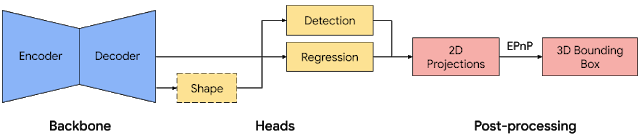}
\end{center}
   \caption{Our network detects the object and estimates the 3D bounding box.}
\label{fig:network}
\end{figure}

Figure \ref{fig:overview} shows an overview of our 3D tracking system. 
Initially, the frames are passed through a single-stage CNN, as shown in Figure~\ref{fig:network}, to predict the object's pose and physical size from a single RGB image. 

\begin{figure}
\begin{center}
   \includegraphics[width=2cm]{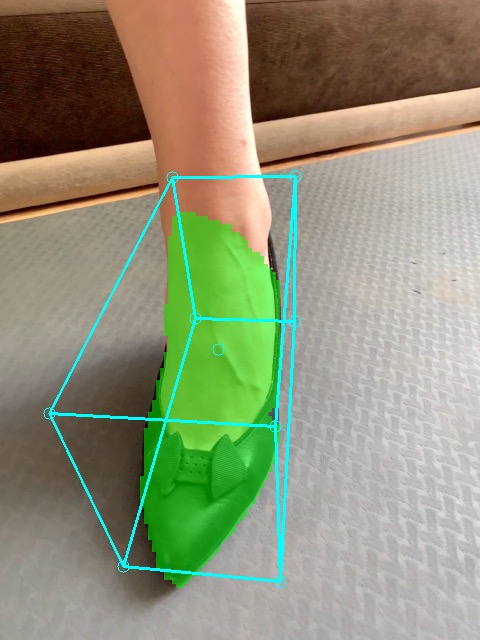}
   \includegraphics[width=2cm]{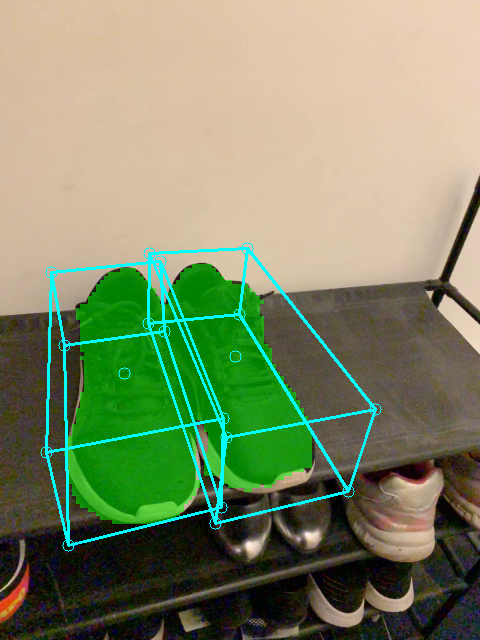}
   \includegraphics[width=2cm]{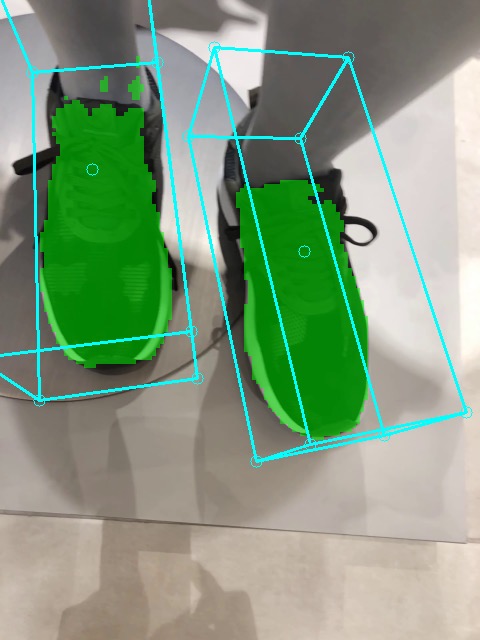}
   \includegraphics[width=2cm]{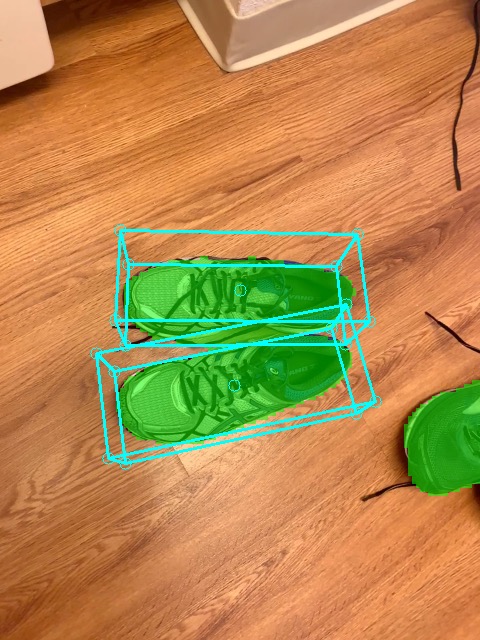}
\end{center}
   \caption{Estimated 3D bounding box and the segmentation mask produced by our object detector network.}
\label{fig:tracking_results}
\end{figure}

The model backbone has an encoder-decoder architecture, built upon MobileNetV2\cite{s2018mobilenetv2}. 
We employ a multi-task learning approach, jointly predicting an object's shape with detection and regression. The shape task predicts the object's shape signals depending on what ground truth annotation is available, \eg segmentation. This is optional if there is no shape annotation present in the training data. For the detection task, we use the annotated bounding boxes and fit a Gaussian to the box, with the center at the box's centroid and standard deviations proportional to the box size. The goal for detection is then to predict this distribution with its peak representing the object’s center location~\cite{ding2019object}. The regression task estimates the 2D projections of the eight bounding box vertices. To obtain the final 3D coordinates for the bounding box, we leverage a well established pose estimation algorithm (EPnP)~\cite{Lepetit_2009_EPnP}. It can recover the 3D bounding box of an object, without a priori knowledge of the object dimensions. Given the 3D bounding box, we can easily compute pose and size of the object. Figure~\ref{fig:network} shows our network architecture and post-processing. The model is light enough to run real-time on mobile devices (at 26 FPS on an Adreno 650 mobile GPU). The details of our model is described in~\cite{Hou:2020}.

After initializing the object's pose with the object detector network, we compute the nine key-points of the bounding box and project them to the image. The nine key-points consists of the bounding box's eight vertices plus its center. We track these points using a planar tracker~\cite{Wei:2019}. Our tracking system consists of a motion analysis module, a robust box tracker, and a pose estimation module for calibration-free 6DoF tracking~\cite{Wei:2019}. Tracking the nine 2D points using a 6-DoF relative scale tracker~\cite{Wei:2019} is sufficient for tracking the object's 9DoF pose. At every frame, we lift the 2D points back to 3D using the EPnP algorithm~\cite{Lepetit_2009_EPnP} to estimate the 3D bounding box (with 9-DoF) up-to scale. Consequently, our model inference only needs to run every few frames, resulting in high efficiency in our mobile pipeline. When a new prediction is made, we consolidate the detection result with the tracking result based on the area of overlap.

\section{Results and applications to AR}
\begin{figure}
\begin{center}
   \includegraphics[width=2.5cm]{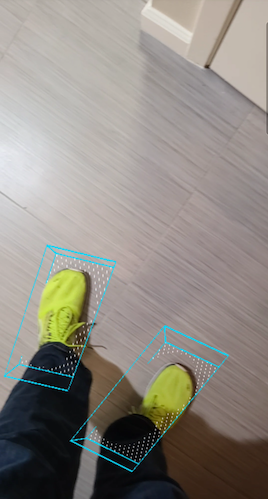}
   \includegraphics[width=2.5cm]{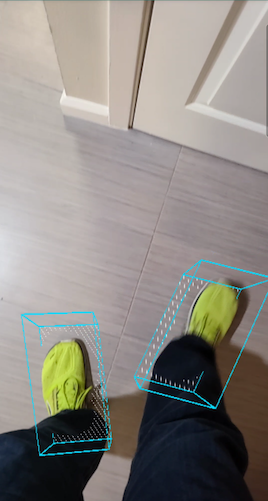}
   \includegraphics[width=2.5cm]{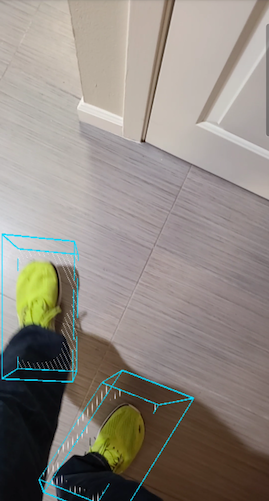}
   \includegraphics[width=2.5cm]{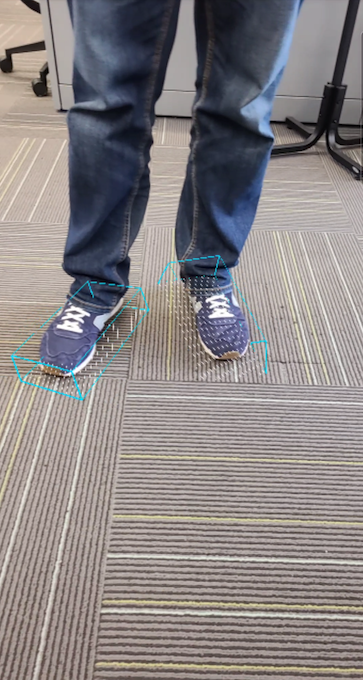}
   \includegraphics[width=2.5cm]{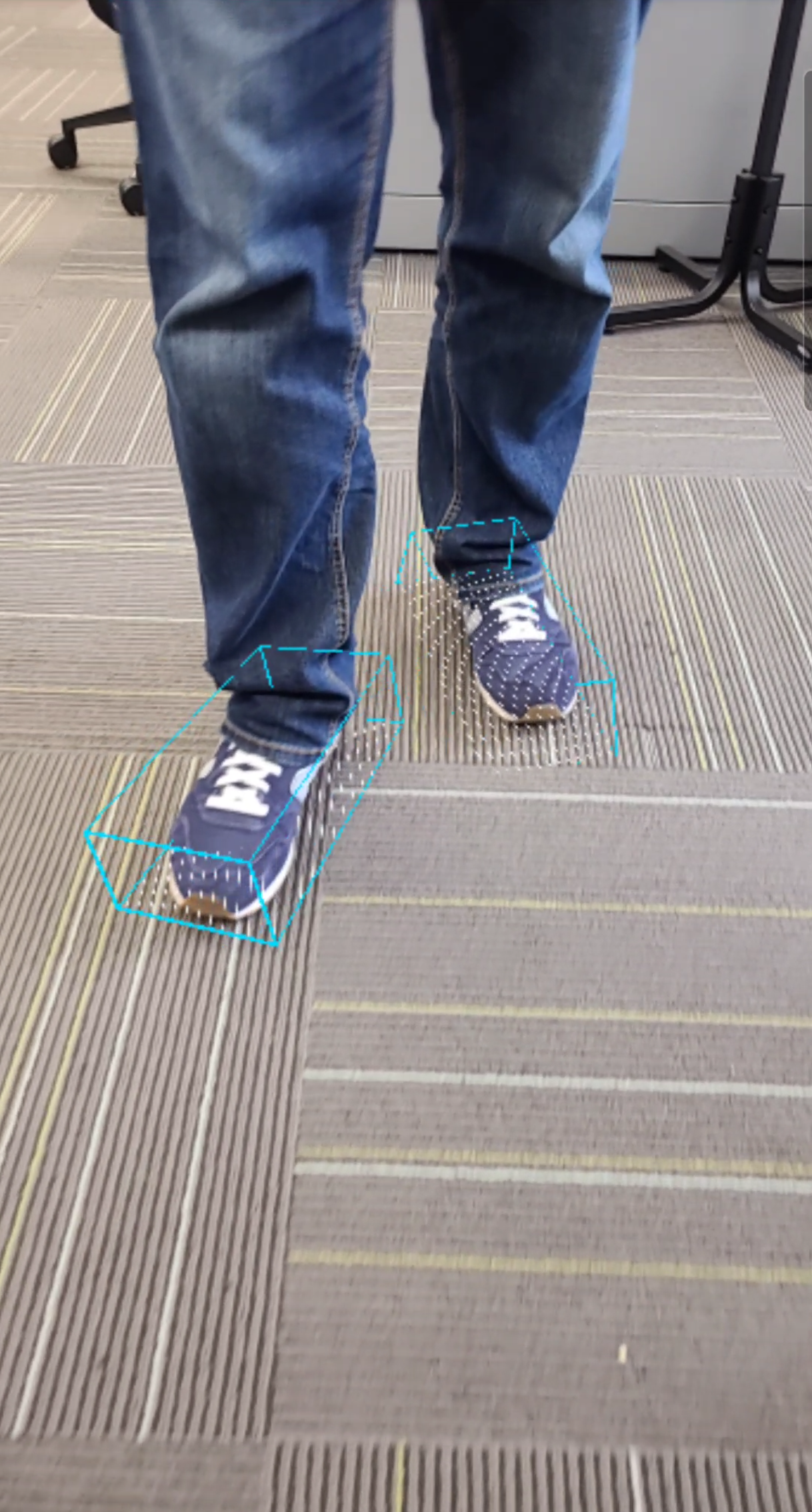}
   \includegraphics[width=2.5cm]{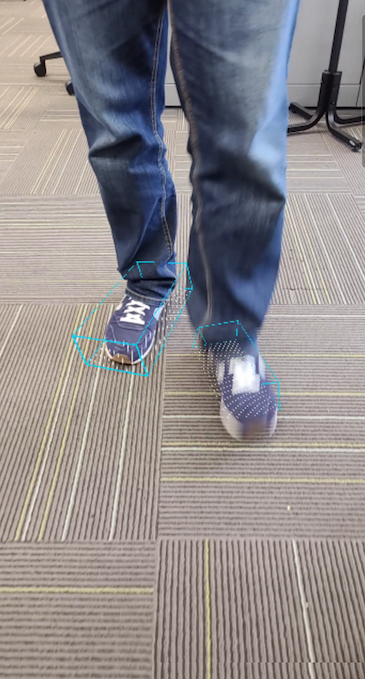}
\end{center}
   \caption{Tracking a 3D bounding box with both camera motion and object movement present.}
\label{fig:tracking_results}
\end{figure}

\begin{figure}
\begin{center}
   \includegraphics[width=3.5cm]{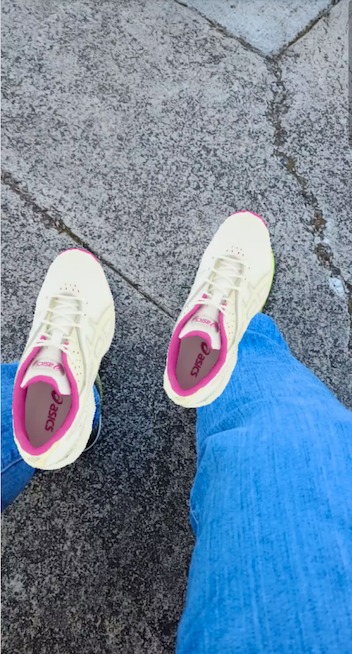}
   \includegraphics[width=3.5cm]{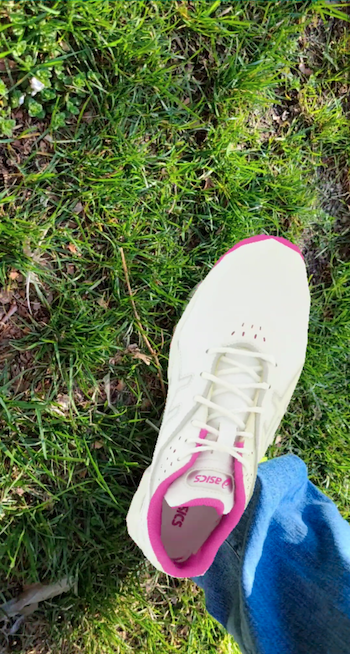}
\end{center}
   \caption{Fitting and rendering a mesh model to the detected objects for AR Applications.}
\label{fig:shoe_render}
\end{figure}

In Figure~\ref{fig:tracking_results}, we demonstrate tracking multiple 3D bounding boxes results from a video. The complete system is implemented in the Mediapipe framework. The model and the code is available at \cite{mediapipe}.
Our detection network predicts the 3D bounding box with average precision of 0.59 at 0.5 3D IoU. The model weights only 5.54MB. The model's output also includes shape information such as segmentation mask. The object detector runs at 26.5fps on the mobile GPU while the 3D tracking runs at 30fps+ on a mobile CPU (Samsung S20 device with Qualcomm's Snapdragon 865 SoC). 

Figure~\ref{fig:shoe_render} shows an example of how to use the tracked object's pose for AR applications such as virtual shoe try-on. In each frame, we render a CAD model at the object pose using OpenGL rendering pipeline. The same polygon mesh model is also rendered in Figure~\ref{fig:tracking_results} inside the bounding box as an occluder to give a 3D effect to the visualization.

We make two key assumptions for the 3D tracking system to work properly: first, there is no gauge ambiguity in the 3D bounding box and the detected eight keypoints are unique. For symmetric objects, e.g. volleyball, this assumption does not hold. As a result, the detection network would predict different orientations each time and that would cause failure when consolidating the detection and tracking results together. The second assumption by the tracking module is that the object's plane does not significantly change while we are tracking it. This assumption is true for most objects without the roll, however, if the tracked object rolls and changes its plane during tracking, e.g. volleyball, we may lose their track.

\section{Conclusion}
In conclusion, we present a system for 3D object tracking that enables real-time instant 3D bounding box tracking on mobile devices. Our proposed system uses a neural network to initialize the 3D pose then utilizes a planar surface tracker to track the object's pose in the video frame. The end-to-end system runs in real-time on mobile devices.

{\small
\bibliographystyle{ieee_fullname}
\bibliography{references.bib}
}

\end{document}